\documentclass[11pt]{article}
\usepackage[preprint]{acl}
\usepackage{times}
\usepackage{latexsym}
\usepackage[T1]{fontenc}
\usepackage{iftex}
\ifPDFTeX
\usepackage[utf8]{inputenc}
\fi
\usepackage{microtype}
\usepackage{inconsolata}
\usepackage{graphicx}
\usepackage{booktabs}
\usepackage{tabularx}
\usepackage{array}
\usepackage{amsmath}
\usepackage{url}

\title{Hybrid Retrieval for Government Scheme Search in English and Marathi}
\author{
Aishwarya Gunjal\textsuperscript{1,2}\quad
Kanchan Khemnar\textsuperscript{1,2}\quad
Pawan Poojary\textsuperscript{1,2}\\
\textbf{Pranav Kulkarni}\textsuperscript{1,2}\quad
\textbf{Raviraj Joshi}\textsuperscript{2}\quad
\textbf{Manish Khodaskar}\textsuperscript{1}\\
\textsuperscript{1}Pune Institute of Computer Technology, Pune, India\\
\textsuperscript{2}L3Cube Labs, Pune, India\\
\small\texttt{\{gunjalaishwarya625, kanchankhemnar170, pawanpoojary989,}\\
\small\texttt{pranavkulkarni1115, ravirajoshi\}@gmail.com};\ \texttt{mrkhodaskar@pict.edu}
}

\begin{document}
\maketitle

\begin{abstract}
Retrieving government schemes requires matching citizens' information needs to formal administrative descriptions, often across languages and terminology. We present a bilingual search system for Maharashtra government schemes and public services that combines BM25 retrieval with sentence-embedding similarity through equal-weight score fusion. The system indexes aligned English--Marathi scheme descriptions and supports multilingual, Marathi-focused, and translation-mediated semantic configurations. Evaluation on manually curated bilingual queries compares BM25, dense retrieval, and hybrid retrieval using IndicSBERT and MahaSBERT. BM25 provides a strong baseline with a mean reciprocal rank (MRR) of 0.864. Hybrid retrieval increases MRR to 0.872 with IndicSBERT and 0.879 with MahaSBERT, whereas the corresponding dense-only systems achieve 0.663 and 0.655. MahaSBERT-based fusion also increases precision from 0.781 to 0.808 while matching BM25 recall at 0.946. These results support the complementary use of lexical and semantic signals in this collection, while showing that improvements over a strong lexical baseline are modest and metric-dependent. The study provides an application-focused analysis of hybrid retrieval for bilingual public-service information access.
\end{abstract}

\section{Introduction}
Government schemes and public services are commonly described using formal names, departmental terminology, and procedural language. Citizens may instead express their needs through short descriptions of a problem or desired benefit. A query such as ``financial help for farmers'' need not share the wording of a scheme described as providing agricultural credit. This vocabulary mismatch motivates retrieval methods that combine exact term matching with semantic representations.

The problem also has a multilingual dimension. A citizen may search in Marathi while relevant descriptions are available in English or in translated administrative language. A useful retrieval system must accommodate this variation without losing the discriminative value of scheme names, categories, and domain-specific terms. Bilingual records provide a common document inventory, but translation alone does not resolve mismatches between a user's phrasing and an official description.

BM25 provides an established lexical ranking function that rewards informative query-term matches \citep{robertson2009bm25}. Sentence embedding models offer a complementary mechanism by representing queries and documents in a shared vector space \citep{reimers2019sentence}. In Marathi, MahaSBERT provides language-focused representations, while IndicSBERT supports multiple Indian languages \citep{joshi2023mahasbert,deode2023l3cube}. However, semantic similarity need not imply that a scheme satisfies a user's specific requirement. Combining lexical and semantic signals therefore requires empirical evaluation against a lexical baseline.

We develop a hybrid search system for a collection of 64 Maharashtra government schemes and public services in English and Marathi. The system combines field-weighted BM25 scores with sentence-embedding similarity and presents ranked records with their source links. We investigate whether fusion improves retrieval over either component and whether the choice between a Marathi-focused and a multilingual encoder affects the resulting ranking.

Our contributions are:
\begin{itemize}
\item A bilingual retrieval pipeline integrating corpus collection, English--Marathi alignment, lexical indexing, and semantic search for government schemes and public services.
\item A comparison of BM25, IndicSBERT-based retrieval, MahaSBERT-based retrieval, and their hybrid combinations on 150 manually curated queries.
\item An analysis of the magnitude and metric dependence of hybrid gains, distinguishing improvements over dense-only retrieval from improvements over an already strong lexical baseline.
\end{itemize}

\section{Related Work}
\paragraph{Lexical and semantic retrieval.}
BM25 estimates relevance from term occurrences, inverse document frequency, and document-length normalization \citep{robertson2009bm25}. It is well suited to exact terminology but does not explicitly represent semantic relationships between different expressions. Contextual encoders such as BERT provide the basis for learned text representations \citep{devlin2019bert}; Sentence-BERT adapts contextual encoding to produce independently computable sentence vectors \citep{reimers2019sentence}. These representations support similarity-based retrieval, although their effectiveness depends on the relationship between the embedding training objective and the target retrieval task. MiniLM uses self-attention distillation to obtain compact Transformer encoders \citep{wang2020minilm}, providing a basis for efficient sentence-embedding configurations. Dense Passage Retrieval (DPR) explicitly trains a dual encoder on question--passage relevance pairs \citep{karpukhin2020dpr}, illustrating the distinction between general sentence similarity and retrieval-specific supervision.

\paragraph{Representations for Indian languages.}
IndicNLPSuite develops corpora, evaluation benchmarks, and pretrained multilingual models for Indian languages \citep{kakwani2020indicnlpsuite}. MuRIL introduces Indian-language pretraining that incorporates translated and transliterated sentence pairs \citep{khanuja2021muril}. More generally, multilingual sentence embedding spaces can be learned through knowledge distillation with parallel sentences \citep{reimers2020multilingual}. LaBSE combines multilingual pretraining with translation-based alignment to learn language-agnostic sentence embeddings \citep{feng2022labse}. MahaSBERT and HindSBERT apply sentence-level training with translated natural language inference and semantic similarity data to Marathi and Hindi \citep{joshi2023mahasbert}. IndicSBERT extends sentence representation learning across multiple Indian languages using multilingual BERT \citep{deode2023l3cube}. These models motivate a comparison between language-focused and multilingual representations for Marathi administrative text. Their performance on general similarity tasks does not, by itself, establish their relative effectiveness for retrieving specific public benefits.

\paragraph{Hybrid score fusion.}
Hybrid retrieval combines lexical and semantic relevance estimates. Reciprocal rank fusion aggregates rank positions rather than raw relevance scores \citep{cormack2009rrf}, providing an alternative to score-level interpolation. \citet{bruch2024fusion} analyze fusion functions, including convex combinations of lexical and semantic scores, and demonstrate the importance of treating score combination as a retrieval-design choice. Our system uses a fixed, equal-weight combination rather than introducing a new fusion algorithm. The contribution lies in its bilingual public-service application and the comparison of encoder choices against a shared lexical baseline.

\paragraph{Multilingual and cross-domain evaluation.}
Mr.~TyDi evaluates dense retrieval in typologically diverse languages and finds that dense signals can improve sparse--dense hybrids even when the dense retriever is weaker than BM25 in isolation \citep{zhang2021mrtydi}. BEIR examines retrieval across heterogeneous tasks and domains and identifies BM25 as a robust zero-shot baseline \citep{thakur2021beir}. Complementary multilingual resources address how retrieval data are constructed: mMARCO translates the MS MARCO passage-ranking dataset \citep{bonifacio2021mmarco}, whereas MIRACL provides native-speaker relevance annotations for monolingual retrieval across diverse languages \citep{zhang2023miracl}. The distinction is relevant to bilingual public-service search, where translated document coverage and native-language relevance judgments serve different purposes. These findings motivate comparison against a strong lexical system and caution against transferring encoder rankings directly from general similarity benchmarks to a specialized collection. Our evaluation addresses a bilingual administrative domain rather than the broader benchmark settings studied in these resources.

\section{Bilingual Scheme Collection}
\subsection{Collection and Preprocessing}
The collection contains 64 distinct government schemes and public services gathered from Maharashtra government portals, including Aaple Sarkar, and scheme information repositories. Each record comprises a title, description, and source URL, together with category information used for retrieval and filtering. The collection covers 14 domain categories, including agriculture, health, housing, education, revenue, transport, welfare, and industry.

We remove HTML markup and navigation content using BeautifulSoup and regular-expression-based cleaning. Text normalization standardizes whitespace and punctuation and corrects encoding artifacts. Potential duplicates are identified through fuzzy matching of names and descriptions; pairs with similarity scores above 85\% are manually reviewed and merged when they refer to the same record. Entries lacking a title or description are corrected from their source or excluded. The resulting scheme identifier is shared across the two language versions.

\subsection{English--Marathi Alignment}
The English records are translated into Marathi using the Google Translation API with domain-specific glossaries. Translation is performed in batches, followed by terminology consistency checks and native-speaker review of a 20\% sample. The aligned collection contains 64 English records and 64 corresponding Marathi records, representing 64 underlying schemes rather than 128 independent schemes.

Table~\ref{tab:corpus} summarizes the collection. Description lengths vary substantially, reflecting differences in the amount of procedural and service information associated with individual records. This variation motivates document-length normalization in lexical retrieval and consistent record construction for semantic indexing.

\begin{table}[t]
\centering
\small
\setlength{\tabcolsep}{4pt}
\begin{tabular}{@{}lrr@{}}
\toprule
Property & English & Marathi \\
\midrule
Records & 64 & 64 \\
Domain categories & 14 & 14 \\
Source groups & 3 & 3 \\
Title length (words) & $3.8\pm2.1$ & $4.6\pm2.4$ \\
Description length (words) & $178.3\pm96.4$ & $171.5\pm92.8$ \\
Description range (words) & 5--620 & 6--590 \\
\bottomrule
\end{tabular}
\caption{Statistics of the aligned collection. Length summaries are expressed as mean $\pm$ standard deviation. Each language version represents the same 64 schemes and services.}
\label{tab:corpus}
\end{table}

\section{Retrieval Framework}
\begin{figure*}[t]
\centering
\includegraphics[width=\textwidth]{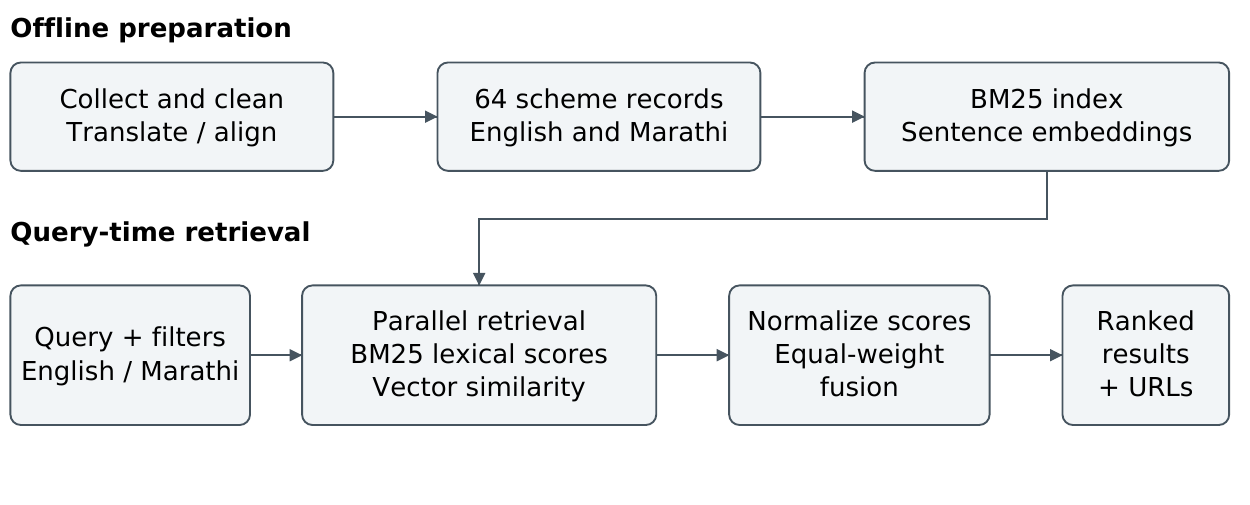}
\caption{Core retrieval pipeline. Offline collection and alignment support lexical and semantic indexes. Query-time rankings are combined through normalized score fusion; the same scheme identifiers link both language versions.}
\label{fig:architecture}
\end{figure*}
\subsection{Architecture}
Figure~\ref{fig:architecture} summarizes the retrieval pipeline. Offline processing produces the aligned records and their lexical and semantic indexes. At query time, the interface accepts an English or Marathi query and optional category, date, or department filters. The backend preprocesses the query, obtains lexical and semantic scores, and combines them to rank the candidate records. Duplicate records are consolidated before presentation. Each result includes its description and source link; keyword highlights and score components provide retrieval-oriented explanations.

\subsection{Lexical Retrieval}
The lexical component uses BM25 in Elasticsearch \citep{robertson2009bm25}, with $k_1=1.5$ and $b=0.75$. Field boosts prioritize scheme titles and category terms relative to descriptions: title, category, and description receive weights of 3, 2, and 1, respectively. This configuration preserves exact scheme-name and domain-term matches while allowing description content to contribute to ranking.

\subsection{Semantic Configurations}
The system supports three semantic configurations:
\begin{itemize}
\item \textbf{IndicSBERT} \citep{deode2023l3cube} produces 768-dimensional representations using a multilingual Indian-language sentence encoder.
\item \textbf{MahaSBERT} \citep{joshi2023mahasbert} produces 768-dimensional representations using a Marathi-focused sentence encoder.
\item \textbf{Translation-mediated MiniLM} \citep{reimers2019sentence,wang2020minilm} maps text into English using Google Translate and applies \texttt{paraphrase-MiniLM-L6-v2}. Translation is an input-processing step rather than a separate embedding model.
\end{itemize}
For an encoder $f$, semantic relevance between query $q$ and record $d$ is estimated using cosine similarity:
\begin{equation}
S_{\mathrm{sem}}(q,d)=\frac{f(q)^{\top}f(d)}{\lVert f(q)\rVert_2\lVert f(d)\rVert_2}.
\end{equation}
Record embeddings are indexed before query processing. The experimental comparison below focuses on IndicSBERT and MahaSBERT, which permit direct analysis of multilingual and Marathi-focused representations under the same retrieval framework.

\subsection{Hybrid Ranking}
Let $\widetilde{S}_{\mathrm{lex}}$ and $\widetilde{S}_{\mathrm{sem}}$ denote normalized lexical and semantic scores. Hybrid ranking uses their convex combination:
\begin{equation}
\begin{split}
S_{\mathrm{hyb}}(q,d)={}&\alpha\widetilde{S}_{\mathrm{lex}}(q,d)\\
&+(1-\alpha)\widetilde{S}_{\mathrm{sem}}(q,d),
\end{split}
\end{equation}
with $\alpha=0.5$. Normalization places the two score streams on comparable scales before fusion. The equal weighting provides a fixed comparison across encoder configurations and does not establish that the weight is optimal for this domain.

Candidates from the lexical and semantic branches are combined and reranked using the hybrid score. The retrieval implementation caps each branch at 2,000 candidates. For this 64-record collection, that cap exceeds the corpus size and therefore imposes no additional depth truncation. The evaluation consequently concerns ranking within a small collection rather than large-scale candidate generation.

\section{Experimental Setup}
\subsection{Queries and Relevance Judgments}
The test set comprises 150 manually curated queries: 75 in English and 75 in Marathi. Each query is associated with one or more relevant schemes in the collection. All configurations are evaluated against the same query-to-scheme relevance mappings. We compare the shared BM25 baseline, the two dense-only retrievers, and the corresponding hybrid configurations.

The query set evaluates retrieval from the fixed collection. It does not measure whether a user meets a scheme's eligibility requirements or whether retrieval leads to a successful application. These outcomes require evidence beyond query--document relevance.

\subsection{Evaluation Metrics}
We use standard relevance-based retrieval measures \citep{manning2008ir}. Mean reciprocal rank (MRR) measures the position of the first relevant result. If $r_q$ is the rank of that result for query $q$, then
\begin{equation}
\mathrm{MRR}=\frac{1}{|Q|}\sum_{q\in Q}\frac{1}{r_q},
\end{equation}
where the reciprocal rank is zero when no relevant result is retrieved.

Precision and recall quantify relevance within the retrieved set. For the returned set $A_q$ and relevant set $G_q$,
\begin{align}
P(q)&=\frac{|A_q\cap G_q|}{|A_q|},\\
R(q)&=\frac{|A_q\cap G_q|}{|G_q|}.
\end{align}
Precision reflects the proportion of returned records that are relevant; recall reflects the proportion of relevant records retrieved. Aggregate precision, recall, and MRR are reported on a 0--1 scale. These metrics capture different aspects of retrieval, so an improvement in one does not imply an improvement in the others.

\section{Results and Discussion}
\subsection{Comparison of Retrieval Strategies}
Table~\ref{tab:performance} reports the retrieval results. BM25 achieves an MRR of 0.864, exceeding IndicSBERT-only retrieval at 0.663 and MahaSBERT-only retrieval at 0.655. Lexical matching is therefore a strong baseline in this collection, and replacing it with dense similarity alone reduces the measured ranking effectiveness.

\begin{table*}[t]
\centering
\small
\setlength{\tabcolsep}{7pt}
\begin{tabular}{@{}llrrr@{}}
\toprule
Retrieval method & Scoring model & Precision & Recall & MRR \\
\midrule
Lexical & BM25 \citep{robertson2009bm25} & 0.781 & 0.946 & 0.864 \\
\midrule
Dense & IndicSBERT \citep{deode2023l3cube} & 0.575 & 0.804 & 0.663 \\
Hybrid & IndicSBERT \citep{deode2023l3cube} & 0.795 & \textbf{0.950} & 0.872 \\
\midrule
Dense & MahaSBERT \citep{joshi2023mahasbert} & 0.562 & 0.775 & 0.655 \\
Hybrid & MahaSBERT \citep{joshi2023mahasbert} & \textbf{0.808} & 0.946 & \textbf{0.879} \\
\bottomrule
\end{tabular}
\caption{Aggregate retrieval results on the bilingual query set. Bold indicates the highest value per metric. The shared lexical baseline is shown once because its scores do not depend on the semantic encoder.}
\label{tab:performance}
\end{table*}

Both hybrid configurations obtain higher precision and MRR than BM25. IndicSBERT-based fusion increases precision by 0.014 and MRR by 0.008, while recall increases from 0.946 to 0.950. MahaSBERT-based fusion increases precision by 0.027 and MRR by 0.015, while recall remains 0.946. Its MRR gain over BM25 is approximately 1.7\% relative to the baseline. These results support a benefit from combining relevance signals, but do not show strict improvement on every metric.

The gains over dense-only retrieval are substantially larger. MahaSBERT-based fusion increases MRR from 0.655 to 0.879, an absolute difference of 0.224, or approximately 34.2\% relative improvement. IndicSBERT-based fusion increases MRR from 0.663 to 0.872, an absolute difference of 0.209. Distinguishing these comparisons is important: the large improvements over dense-only retrieval do not characterize the smaller improvement over BM25.

\subsection{Effect of the Semantic Encoder}
IndicSBERT obtains higher precision, recall, and MRR than MahaSBERT in the dense-only setting. Under hybrid fusion, MahaSBERT instead yields the highest precision and MRR, while IndicSBERT yields the highest recall. The encoder ranking therefore depends on the retrieval configuration and evaluation metric.

A possible explanation is that an encoder's usefulness in fusion depends on which candidates it promotes relative to BM25, rather than only on its standalone ranking. The results are compatible with complementary relevance signals, but do not isolate the mechanism responsible for the change in ordering. They also do not establish language-specific superiority from the aggregate bilingual scores.

\subsection{Implications for Public-Service Search}
Administrative descriptions contain names, categories, and terms that can make lexical matching effective. Semantic representations may contribute when users paraphrase those descriptions, but can also assign high similarity to schemes with related topics and different purposes. The observed results favor retaining lexical evidence while supplementing it with an embedding-based signal. This qualitative pattern is consistent with the complementary sparse and dense signals observed in Mr.~TyDi \citep{zhang2021mrtydi}, although the collections and evaluation settings differ.

Qualitative error analysis identifies ambiguous queries and semantically similar schemes across departments as challenging cases. A retrieved record can be topically related without satisfying a user's intended benefit or eligibility condition. Source links and explicit scheme details therefore remain important for interpreting results. The system supports discovery of potentially relevant information rather than automated eligibility determination.

\section{Conclusion}
We presented a bilingual hybrid retrieval system for 64 Maharashtra government schemes and public services. The system combines field-weighted BM25 with sentence-embedding similarity through equal-weight normalized score fusion. On 150 English and Marathi queries, MahaSBERT-based hybrid retrieval achieves the highest precision (0.808) and MRR (0.879), while IndicSBERT-based hybrid retrieval achieves the highest recall (0.950). BM25 remains a strong baseline, with MRR 0.864, and substantially exceeds both dense-only systems.

The main finding is that hybrid retrieval provides modest, metric-dependent improvements over lexical retrieval in this collection, rather than a uniform advantage on all measures. Future work should evaluate larger and more diverse scheme collections, domain-specific representation learning, and query-dependent fusion weights, alongside direct assessment of user information needs and eligibility-sensitive retrieval.

\section*{Limitations}
\paragraph{Collection and query coverage.}
The evaluation contains only 64 schemes and services and 150 manually curated queries. This scale does not establish performance across the full range of government programs, departments, or citizen language use. Bilingual aggregate results also do not establish the relative performance of the encoders separately for English and Marathi.

\paragraph{Translation and domain terminology.}
The Marathi descriptions are translated from English. Native-speaker review of a sample mitigates some translation errors but does not ensure complete semantic equivalence. Administrative terminology and eligibility conditions may not be represented adequately by general sentence embeddings.

\paragraph{Retrieval scope and generalization.}
The fixed fusion weight and small corpus do not establish optimal weighting or scalability. The system indexes textual records rather than scanned forms or other multimodal documents. Periodic corpus updates can also delay coverage of newly introduced or modified schemes. Evaluation concerns relevance ranking and does not demonstrate improvements in public awareness, benefit uptake, or user eligibility decisions.

\section*{Ethical Considerations}
Government scheme descriptions may change, and incorrect or outdated results can mislead users seeking public support. Retrieved records should retain links to authoritative sources, and users should consult current official eligibility and application information. Translation and ranking errors may disproportionately affect users relying on Marathi descriptions. Any future use of personal information for eligibility-aware retrieval would require appropriate consent, data minimization, and privacy safeguards.

\begin{samepage}
\section*{Acknowledgements}
This work was carried out under the mentorship of L3Cube, Pune. We would like to express our gratitude towards our mentor for his continuous support and encouragement. This work is a part of the L3Cube-MahaNLP project \citep{joshi2022mahanlp}.
\end{samepage}

\bibliography{main}
\end{document}